\documentclass{article}

    \PassOptionsToPackage{numbers, compress}{natbib}

    \usepackage[preprint]{neurips_2020}

\usepackage[utf8]{inputenc} 
\usepackage[T1]{fontenc}    
\usepackage{hyperref}       
\usepackage{url}            
\usepackage{booktabs}       
\usepackage{amsfonts}       
\usepackage{nicefrac}       
\usepackage{microtype}      
\usepackage{graphicx}
\usepackage[misc]{ifsym}

\usepackage{amsmath}
\usepackage{amssymb}
\usepackage{multirow}
\usepackage{makecell}
\usepackage{threeparttable}
\usepackage{lipsum}
\usepackage{tabularx}
\usepackage{diagbox}
\usepackage{color}
\usepackage{enumerate}

\title{POS-BERT: Point Cloud One-Stage BERT Pre-Training}

\author{Kexue Fu$^{1,2,3}$ \ \ Peng Gao$^{2}$ \ \ ShaoLei Liu$^{1,3}$ \ \ Renrui Zhang$^{2}$ \\
\textbf{Yu Qiao}$^{2}$ \ \ \textbf{Manning Wang}$^{1,3}$\thanks{Corresponding author}\\
1 Digital Medical Research Center, School of Basic Medical Sciences, Fudan University \\
2 Shanghai AI Lab \\
3 Shanghai Key Laboratory of Medical Image Computing and Computer Assisted Intervention\\
{\tt $\{$fukexue, mnwang$\}$@fudan.edu.cn, gaopeng@pjlab.org.cn}}

\begin{document}

\maketitle

\begin{abstract}
  Recently, the pre-training paradigm combining Transformer and masked language modeling has achieved tremendous success in NLP, images, and point clouds, such as BERT. However, directly extending BERT from NLP to point clouds requires training a fixed discrete Variational AutoEncoder (dVAE) before pre-training, which results in a complex two-stage method called Point-BERT. Inspired by BERT and MoCo, we propose POS-BERT, a one-stage BERT pre-training method for point clouds. Specifically, we use the mask patch modeling (MPM) task to perform point cloud pre-training, which aims to recover masked patches information under the supervision of the corresponding tokenizer output. Unlike Point-BERT, its tokenizer is extra-trained and frozen. We propose to use the dynamically updated momentum encoder as the tokenizer, which is updated and outputs the dynamic supervision signal along with the training process. Further, in order to learn high-level semantic representation, we combine contrastive learning to maximize the class token consistency between different transformation point clouds. Extensive experiments have demonstrated that POS-BERT can extract high-quality pre-training features and promote downstream tasks to improve performance. Using the pre-training model without any fine-tuning to extract features and train linear SVM on ModelNet40, POS-BERT achieves the state-of-the-art classification accuracy, which exceeds Point-BERT by 3.5\%. In addition, our approach has significantly improved many downstream tasks, such as fine-tuned classification, few-shot classification, part segmentation. The code and trained-models will be available at: \url{https://github.com/fukexue/POS-BERT}.
\end{abstract}

\section{Introduction}
\label{sec:intro}
Point cloud is an intuitive, flexible and memory-efficient 3D data representation and has become indispensable in 3D vision. Learning powerful point cloud representation is very crucial for facilitating machines to understand the 3D world, which is beneficial for promoting the development of many important real-world applications, such as autonomous driving \cite{ref28}, augmented reality \cite{ref30} and robotics \cite{ref29}. With the rapid development of deep learning in these years \cite{ref43,ref44}, supervised 3D point cloud analysis methods have made great progress \cite{ref19,ref20,ref21,a35}. However, both exponentially increasing demand for data and expensive 3D data annotation hinder further performance improvement of supervised methods. On the contrary, due to the widespread popularity of 3D sensors (Lidar, ToF camera, RGB-D sensor or camera stereo-pair), a large number of unlabeled point cloud data are available for self-supervised point cloud representation learning.

Unsupervised or self-supervised learning methods have shown their effectiveness in different fields \cite{ref22,ref23,ref24,ref26,ref27}. Recent work \cite{method3, ref23, method2, ref26, method11} has achieved good performance by combining point clouds with self-supervised learning techniques, such as generative adversarial networks (GAN) \cite{ref23}, variational autoencoders (VAE) \cite{ref22}, and Gaussian mixture models (GMM) \cite{ref24}. These methods usually rely on tasks such as distribution estimation or reconstruction to provide supervisory signals, and can learn good local detail features, but it is difficult to capture higher-level semantic features. To learn higher-level semantic features, some methods learn point cloud representations, such as orientation estimation, by constructing a series of transformation prediction tasks \cite{a10, a11, a12}. Inspired by unsupervised learning of 2D images \cite{ref12, ref13, ref17}, point cloud representation is learned by constructing a series of contrast views \cite{a14, a15, a16, a17} and combining the most advanced comparative learning methods. However, these methods rely on network structures with specific inductive bias to achieve good performance, such as PointNet++, DGCNN, and so on. In addition, previous methods have never studied the performance of standard transformers in point cloud analysis tasks.

\begin{figure*}[ht]
  \centering
  \includegraphics[width=\textwidth]{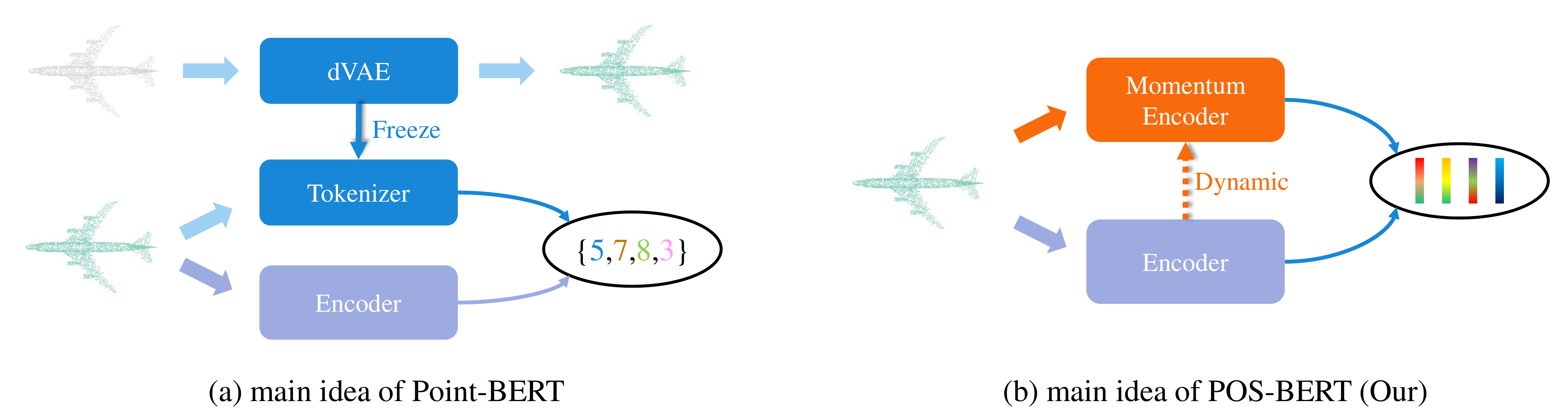} 
  \caption{The difference between our approach and Point-BERT. (a) Point-BERT uses an additional pre-trained dVAE as the Tokenizer, which is frozen during training and the output is discrete. (b) Our approach eliminates the need for extra processing stages, and Tokenizer is derived from Encoder through momentum updates, which are dynamic during the training process and the output is continuous.}\label{fig1}
\end{figure*}

Recently, Transformer has achieved impressive results in language and image tasks through extensive unlabeled data learning and is becoming increasingly popular. Inspired by NLP, Point-BERT devise a mask patch modeling (MPM) task to pre-train point cloud Transformers. To generate meaningful representations for masked patches to guide point cloud Transformers learning, Point-BERT additionally trains a discrete Variational AutoEncoder (dVAE) based on DGCNN as a tokenizer, as shown in Fig.\ref{fig1} (a). As a result, Point-BERT is a two-stage approach, in which the weight of tokenizer is frozen, and its feature extraction capabilities directly affect the learning of point cloud Transformers. Unlike Point-BERT, we extract meaningful representations of masked patches by replacing the frozen tokenizer with momentum encoder, which is dynamically updated, as shown in Fig.\ref{fig1} (b). Therefore, our approach is one-stage, and the meaningful representation of mask patches will become better as the training progresses.
In this article, we propose a one-stage BERT point cloud pre-training method named POS-BERT. Inspired by BERT and MoCo, we used MPM task to pre-train on point cloud and chose standard Transformer without specific inductive biases as backbone. Specifically, we first divide the point cloud into a series of patches, then randomly mask out some patches and feed them into an encoder based on standard transformer. Then, we use a dynamically updated momentum encoder as the tokenizer. The Momentum Encoder has the same network structure as the Encoder, but it does not have gradient backward. Its weight is jointly optimized with MPM through momentum update during the pre-training stage. This greatly simplifies the pre-training step. Next, the point cloud patches before masked are fed to the Momentum Encoder. The objective of MPM is to make the Encoder recover output consistent with the Momentum Encoder output at the masked patches position as much as possible. However, recovering the masked patch information independently leads to limited ability of point cloud transformer's class token to extract high-level semantic information. To address this problem, we perform contrastive learning to maximize the class token consistency between different augmentation (for example, cropping) point cloud pairs. The main contributions are summarized as follows:
\begin{enumerate}[1)]
  \item We propose a Point Cloud One-Stage BERT pre-training method, and named POS-BERT. We use momentum encoder to provide continuous and dynamic supervision signals for masked patches in mask patch modeling pretext task. The Momentum Encoder is updated dynamically during the pre-training stage and does not require extra pre-training processing.
  \item We introduce a contrastive learning strategy on transformer's class token between different augmentation point cloud pairs, which can help point cloud transformer's class token obtain a better high-level semantic representation.
  \item Experiments demonstrate that POS-BERT achieves state-of-the-art performance in linear SVM classification task and downstream tasks, such as classification and segmentation.
\end{enumerate}

\section{Related work}
\textbf{Point Cloud Self-Supervised Learning} The goal of self-supervised learning is to learn good feature representations from unlabeled raw data so that they can be well adapted to various downstream tasks \cite{a1}. Currently, self-supervised learning has been extensively studied in point cloud representation learning, and they focus on constructing a pretext task to help the network better learn 3D point cloud representations. A commonly adopted pretext task is to reconstruct the input point cloud from the latent encoding space, which can be implemented through Variational AutoEncoders \cite{a2,a3,a4,a5,a6,a13}, Generative Adversarial Learning (GANs) \cite{a7,a8}, Gaussian Mixed Model \cite{ref24,a9}, etc. However, these methods are computationally expensive, and rely excessively on reconstructing local details, making it difficult to learn high-level semantic features. Hence, some researchers employed Transformation Prediction as a prediction pseudo-task. Sauder et al. \cite{a10} proposed to use jigsaw puzzle as a pretext task for 3D point cloud representation learning. Wang et al. \cite{a11} destroyed the point cloud and then pretrained the network by a self-supervised manner with the help of point cloud complementation task. Poursaeed et al. \cite{a12} used orientation estimation as a pretext task by randomly rotating the point cloud and then allowing the network to predict the rotation. As contrastive learning becomes increasingly popular, Jing and Afham et al. \cite{a14, a15}, proposed a task training network for finding cross-modality correspondences. Specifically, they obtain the corresponding 2D view by rendering the 3D model, and then extracts 2D view features and 3D point cloud features using 2D convolutional networks and graph convolutional networks. Finally, the instance correspondence between the two modalities is estimated based on these features. Qi et al. \cite{a19} calculated the contrastive loss on matched point pairs by rigidly transforming the point clouds with feature vectors for each point of the two point clouds before and after the transformation. Wang et al. \cite{a16} designed a multi-resolution contrastive learning training strategy that can train point-by-point and shape feature vectors simultaneously. Inspired by BYOL \cite{a18}, Huang et al. \cite{a17} constructed point cloud pairs that undergo spatio-temporal transformations, and forced the network to learn the consistency between different augmented views. However, all previous studies resort to point cloud domain-specific network architectures to achieve promising performance, which would greatly hinder the development of deep learning towards a generalized model. 
More importantly, these studies have never investigated self-supervised representation learning using a transformer-based point cloud processing network. Recently, Point-BERT \cite{a20} has proposed a modeling approach using standard transformer network combined with mask language modeling for the first time to achieve self-supervised representation learning of point clouds, which is a direct extension of BERT \cite{a27} (popular in the field of NLP) on point clouds. However, there is no mature BPE \cite{a26} algorithm in the point cloud domain as in NLP, leading to a lack of an effective vocabulary to guide the learning of mask language modeling. For this reason, Point-BERT \cite{a20} pre-trained a discrete Variational AutoEncoder (dVAE) \cite{a21} as tokenizer through an additional point cloud network DGCNN to construct vocabularies for point cloud patches. This directly brings about two problems: First, the whole method becomes a complex two-stage solution; Second, the weights of the pre-trained tokenizer are frozen and cannot change adaptively with the network training process, and the performance of the fixed tokenizer will directly doom the performance of the pre-trained model. Unlike Point-BERT, we use dynamically updated momentum encoder instead of a frozen tokenizer to extract features from point cloud patches. Additionally, our solution is one-stage, and the Momentum Encoder can be continuously updated as the network training progresses, providing the network with a suitable feature representation of point cloud patches for the current training stage.

\textbf{Transformers} Transformer has made great advances in the field of machine translation and natural language processing with its long-range modeling capability brought by the attention mechanism. Inspired by the successful applications of Transformer in NLP field, it has also been introduced into the image field \cite{a29, a30, a34}, leading to  backbone networks such as ViT \cite{a29}, SWin \cite{a30}, Container\cite{ref44}, etc., which surpassed CNN-based ResNet and showed excellent performance in downstream tasks such as classification \cite{a29}, segmentation \cite{a32}, object detection \cite{a33}. Although there is a trend of grand unification of transformer in the field of NLP and image, the development of transformer in the field of point cloud is highly slow. PCT \cite{ref35} and PointTransformer \cite{a31} have modified the transformer layer in standard transformer and combined with layer aggregation operation to achieve point cloud classification and segmentation. Unlike these approaches, Point-BERT \cite{a20} achieves comparable performance with a standard transformer without introducing a bias structure, but it requires a specific point cloud network DGCNN to provide supervised signals for pre-training. By comparison, our proposed method completely rejects the introduction of other networks and uses only the standard transformer-based network to learn point cloud representations.

\textbf{Mask Language Modeling Paradigm} Mask language modeling was proposed in BERT \cite{a22}, which revolutionized the pre-training paradigm for natural language. Inspired by BERT, Bao et al. proposed BEiT \cite{a23} for pre-training a standard transformer applicable to images. It maps the input image patches into meaningful discrete tokens by dVAE \cite{a21}, then randomly masks some of the image patches, and feeds the masked image patches and the remaining images into the standard transformer to reconstruct the tokens of these masked image patches. Following BEiT, Zhou et al. \cite{a35} perform masked prediction with an online tokenizer. Unlike BEiT, He et al. \cite{a24} trained the network by directly reconstructing the original image patches. Inspired by BEiT, Yu et al. \cite{a20} proposed Point-BERT for point cloud pre-training and demonstrated that the MLM paradigm is feasible for point cloud pre-training. We inherit the idea of Yu et al. and also adopt the MLM approach for point cloud pre-training.

\textbf{Contrastive learning} Contrastive learning is a branch of self-supervised learning, which learns knowledge from the data itself without the demand of data annotation. The main idea of contrastive learning is to maximize the consistency between positive sample pairs and the differences between negative sample pairs. Representative methods of contrastive learning include MoCo series \cite{ref11,ref12,ref13} and SimCLR \cite{ref14}. Recently, BYOL \cite{ref17} and Barlow Twins \cite{ref18} pointed out that only using positive samples can still obtain powerful features. In this paper, we introduce the idea of contrastive learning to help point cloud Transformer learn the high-level semantic representation.

\section{Method}
\begin{figure*}[ht]
  \centering
  \includegraphics[width=\textwidth]{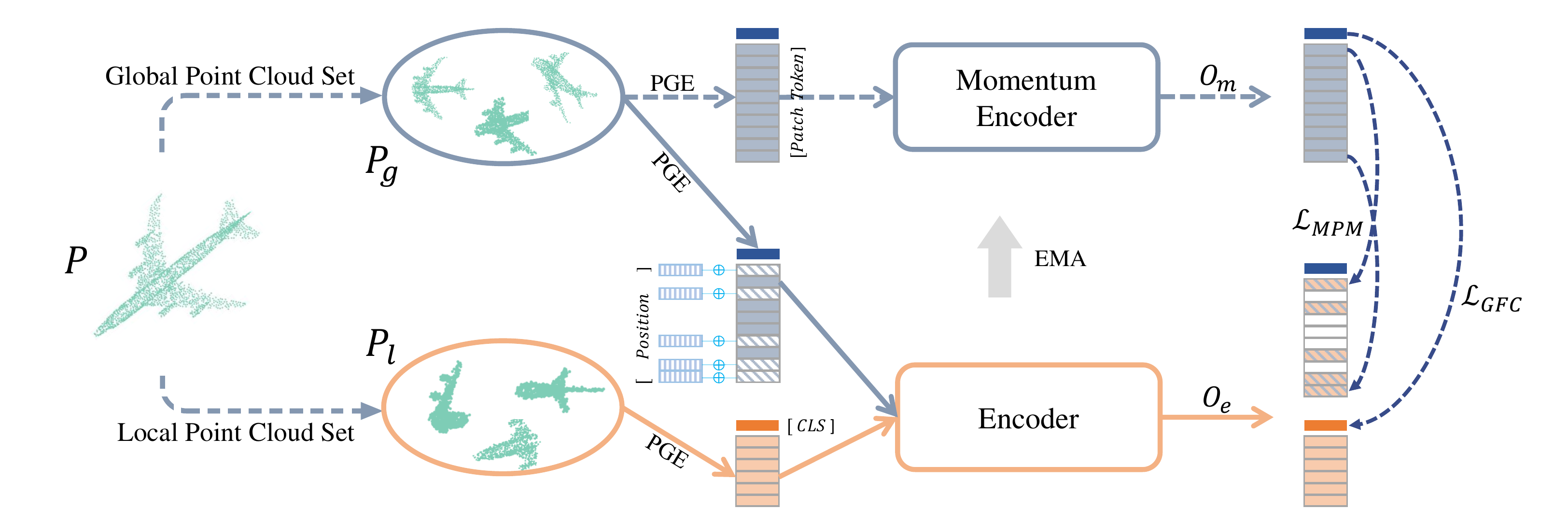} 
  \caption{The overall framework of POS-BERT. PGE represents patch generation and embedding module, CLS represents class token, EMA represents exponential moving average, solid line represents gradient back-propagation, and dotted line represents stop-gradient operator.}
  \label{fig2}
\end{figure*}

We propose a Point Cloud One-Stage BERT pre-training approach \textbf{POS-BERT}, which is simple and efficient. Fig.\ref{fig2} illustrates the overall framework of POS-BERT. Firstly, the global point cloud set $P_{g}$ and the local point cloud set $P_{l}$ are obtained by cropping the raw point clouds $P \in R^{N \times 3}$ with different cropping ratios. Then, we use the PGE module to divide both global and local point clouds into smaller patches with fixed number of points and embed the patches into high-dimensional representation (patch token) though standard Transformer-based encoders. Because local point clouds do not represent complete objects very well, only global point clouds are input into the Momentum Encoder, which is dynamically updated to encode meaningful representations to provide learning objectives for the Encoder. The Encoder is trained using the mask patch modeling task to match the Momentum Encoder outputs. Some patches of the global point clouds are randomly masked out and position information is added to the corresponding masked patches, and then they are input into the Encoder together with the local point cloud set. Finally, we calculate the mask patch modeling loss $\mathcal{L}_{MPM}$ between the Encoder outputs' patch tokens and the Momentum Encoder outputs' patch token, and the global feature loss loss $\mathcal{L}_{GFC}$ between the Encoder outputs' class token and the Momentum Encoder outputs' class token. Overall, our framework consists of four key components: Encoder, Momentum Encoder, Mask Patch Modeling and Loss Function and they will be introduced in detail the following part of this section. We will start with Section \ref{sec31} on how to transform point into patch embedding with the Encoder. Next, mask patch modeling is described in section \ref{sec32}. Then we introduce the dynamic tokenizer implemented by the Momentum Encoder for providing supervision for the MPM tasks in section\ref{sec33}. Finally, we describe our loss function in section \ref{sec34}.

\subsection{Point2Patch Embedding and Encoder Architecture} \label{sec31}

\begin{figure*}[ht]
  \centering
  \includegraphics[width=\textwidth/2]{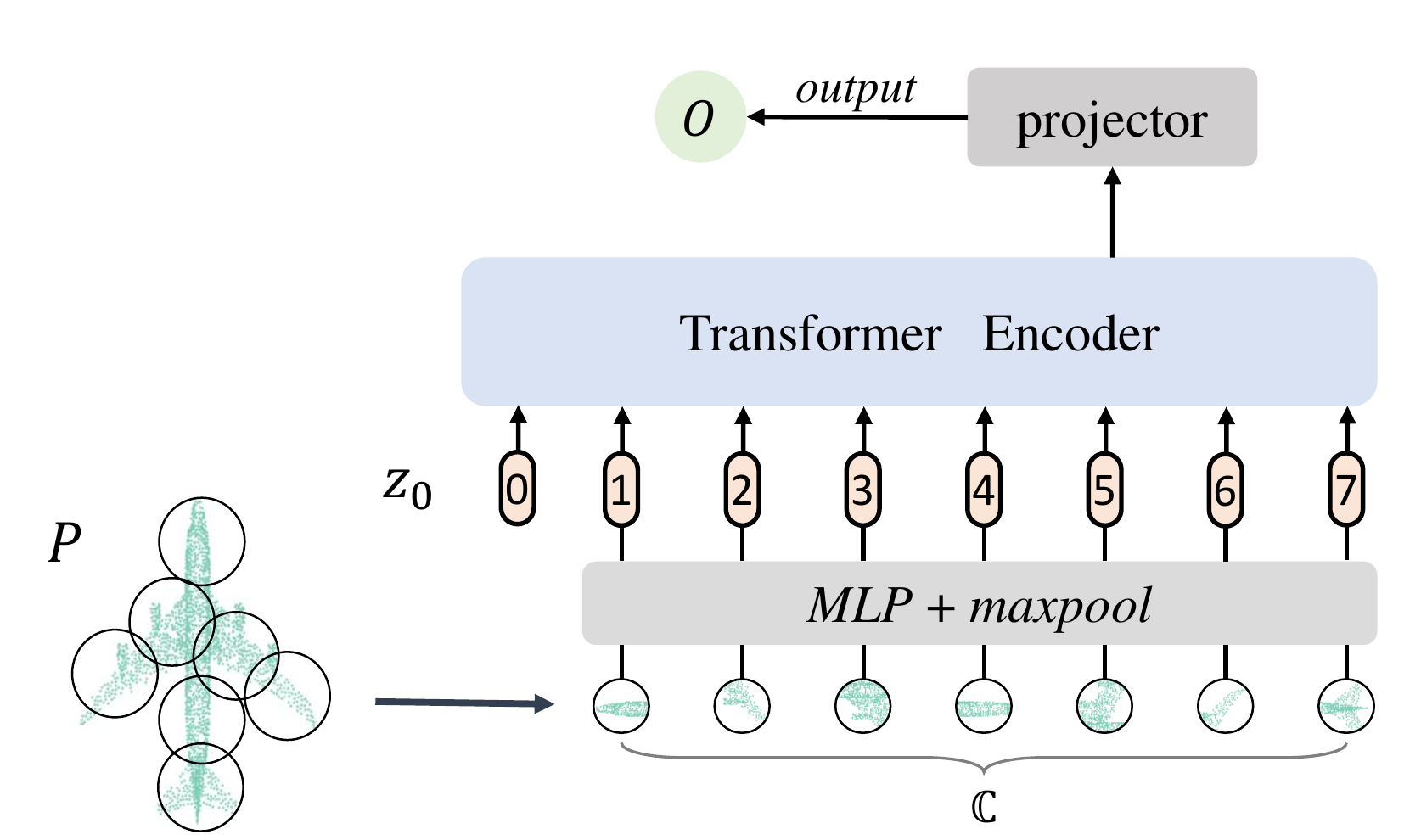} 
  \caption{The architecture of standard transformer-based Encoder.}
  \label{fig3}
\end{figure*}

The simplest way to extract point cloud features is to input each point into the transformer as one token. Because the complexity of transformer is $O\left(N^{2}\right)$, where $N$ is the length of the input token, extracting feature of each point directly will result in memory explosion. Fig.\ref{fig3} describes the overall pipeline of the Transformer-based feature extraction in this paper. Following Point-BERT, we divide a given global/local point cloud $P$ into local patches with a fixed number of $K$ points. In order to minimize overlap between patches $\left\{p_{i} \in \mathbb{C} \mid i=1 \ldots Q\right\}$, we first calculate the number of patches $Q=\operatorname{ceil}(N / K)$, then use farthest point sampling (FPS) algorithm to sample the center point $c_{i}$ of each patch. The $k$-nearest neighbor algorithm is used to obtain $K$ neighbors for each center point, and the center point and corresponding neighbor points form a local patch $p_{i}$. Next, Using the PointNet and maxpooling operations to map point coordinates of each patch to a high-dimensional embedding as patch tokens. Finally, these patch tokens are fed into the standard transformer with a learnable class token. 

We used a standard transformer as the Encoder backbone, which consists of a series of stacked multi-head self-attention layers and fully connected feed-forward network. As mentioned earlier, class tokens $\left\{t_{0}\right\}$ and a series of patch tokens $\left\{t_{1}, \ldots, t_{M}\right\}$ are concatenated along the patch dimension to get the transformer's input $T_{0}=\left\{t_{0}, t_{1}, \ldots, t_{Q}\right\} \in R^{(Q+1) \times D}$. After $T_{0}$ passes through the h-layer transformer block, we get the feature of each patch $T_{\mathrm{h}}=\left\{t_{0}^{h}, t_{1}^{h}, \ldots, t_{Q}^{h}\right\}$ with global receptive field. Finally, we map the features of each patch to the loss space, where the projector is composed of multiple layers of $MLP$. In the inference stage and downstream tasks, we do not need the projector. Decoupling the feature representation and loss function can make the learned patch's features more general.

\subsection{Mask Patch Modeling} \label{sec32}
Inspired by Point-Bert, we also use a mask patch modeling task to pretrain the point cloud Transformer. As described in Section \ref{sec31}, we have obtained the transformer's input $T_{0}=\left\{t_{0}, t_{1}, \ldots, t_{M}\right\}$. Masked patch tokens $\mathcal{M}$ is obtained by randomly masking the tokens of some patches in $T_{0}$, except $t_{0}$. Next, we randomly mask/replace [20\%, 40\%] patch tokens with a learnable mask token $E[m] \in R^{D}$, where masked tokens are defined as $m_{t}$. Then, the center point position embedding $\operatorname{pos}=\operatorname{mlp}\left(c_{i}\right)$ corresponding to patch tokens is added to $m_{t}$, $c_{i}$ represents the $xyz$ coordinate of the patch center point. Finally, the transformer's input tokens obtained by high-dimensional embedding after masking can be expressed as $\widehat{T}_{0}=\left\{t_{0}\right\} \cup\left\{t_{i} \mid i \notin \mathcal{M}\right\}_{i=1}^{Q} \cup\left\{E[m]+\operatorname{pos}_{i} \mid i \in \mathcal{M}\right\}_{i=1}^{Q}$, and the lost information of masked tokens is recovered from $\widehat{T}_{0}$ through Encoder.

\subsection{Dynamic Tokenizer by Momentum Encoder} \label{sec33}
Momentum Encoder is often used in contrastive learning to provide a global semantic supervision for target network. Inspired by MoCo, we propose a dynamically updated tokenizer, which is implemented by momentum Encoder. Grill's preliminary experiments show that even using the output of random initialization network as supervision, target network can also learn a better output representation than random initialization network \cite{ref17}. This result provides a strong support for the replacement of dVAE by the dynamically updated momentum encoder during early training. Therefore, we use a random network to initialize the Momentum Encoder. Although randomly initialized networks can help Encoder get better representation in the early stages of training, if the performance of tokenizer is not continuously improved, the ability of Encoder will stop as tokenizer stops. Accordingly, we need a tokenizer that can dynamically update and improve its quality while at the same time its output does not change rapidly before and after each update. The momentum encoder in contrastive learning solves these two concerns well, and its update formula is as follows:

\begin{equation}
  \theta_{m}=\lambda \theta_{m}+(1-\lambda) \theta_{e}
\end{equation}

where, $\theta_{m}$ represents the weight of Momentum Encoder, $\theta_{e}$ represents the weight of Encoder. $\lambda \in[0,1)$ is a momentum coefficient, which follows a cosine schedule from 0.996 to 1 during training. 

Momentum Encoder enhances itself by constantly introducing new knowledge learned from Encoder, so Momentum Encoder also has the ability to recover lost information. Moreover, it dynamically integrates the Encoder weights of multiple training stages, and has better feature extraction ability than the Encoder. Therefore, our final pre-training model weights come from Momentum Encoder.

\subsection{Loss Function} \label{sec34}
We hope that the pre-training model can not only recover the lost information, but also learn the high-level semantic representation. Therefore, our loss function consists of two parts: mask patch modeling loss  $\mathcal{L}_{MPM}$ and global feature contrastive loss $\mathcal{L}_{GFC}$.

For mask patch model loss $\mathcal{L}_{MPM}$, we encourage the Encoder to recover the information lost by masked patch under the supervision of meaningful representations, which is generated by Momentum Encoder. The formula of mask patch model loss is as follows:
\begin{equation}
  \mathcal{L}_{M P M}=\min _{\theta_{e}} \sum_{i \notin \mathcal{M}}-O_{m}^{i} \cdot \log \left(O_{e}^{i}\right)
\end{equation}

where, $O_{m}^{i}$ represents the output of the Momentum Encoder corresponding to the $i$-th patch, $O_{e}^{i}$ represents the output of the Encoder corresponding to the $i$-th patch.

Although the idea of contrastive learning was also used in Point-BERT to achieve high-level semantic features, the results were not ideal, which can be observed from Tab .\ref{tab1}. In addition, it needs to maintain a memory bank to store a large number of negative samples, which takes up a large amount of storage space. In contrast, we utilize different cropping rate to obtain different augmentation state point clouds: global point clouds and local point clouds with the following formula:
\begin{equation}
  \begin{aligned}
  &P_{g}^{i}=\operatorname{crop}\left(P, \operatorname{rand}\left(r_{g 1}, r_{g 2}\right)\right), \quad i=1 \cdots I \\
  &P_{l}^{j}=\operatorname{crop}\left(P, \operatorname{rand}\left(r_{l 1}, r_{l 2}\right)\right), \quad j=1 \cdots J
  \end{aligned}
\end{equation}

where $\operatorname{crop}(\cdot\ , \cdot)$ represents cropping an area at a fixed ratio, represented by the second parameter. $\operatorname{rand}(\cdot\ , \cdot)$ generates a random value between the maximum and the minimum values. Here, $r_{g1}$ and $r_{g2}$ are the minimum and maximum cropping ratio for generating the global point cloud set, respectively. Similarly, $r_{l1}$ and $r_{l2}$ are the minimum and maximum cropping ratios for generating of the local point cloud set, respectively. $I$ and $J$ are the number of point clouds in $P_{g}$ and $P_{l}$, respectively. During training phase, the Encoder encodes masked global point clouds and local point clouds, while the Momentum Encoder only encodes global point clouds.

\begin{equation}
  \mathcal{L}_{GFC}=\min _{\theta_{e}} \sum_{i=1}^{I} \sum_{j \neq i}^{J}-\left(O_{m}^{c l s}\right)_{i} \cdot \log \left(\left(O_{e}^{c l s}\right)_{j}\right)
\end{equation}

Finally, we combine all the above-mentioned loss function as our final self-supervised objectives:

\begin{equation}
  \mathcal{L}=\omega_{1} * \mathcal{L}_{M P M}+\omega_{2} * \mathcal{L}_{GFC}
\end{equation}

where, the hyperparameters $\omega$ control the balance between loss functions, for all the experiments in this paper, we set $\omega_{1}=0.5$, $\omega_{2}=1.0$.

\section{Implementation and Dataset}
\subsection{Implementation}
\textbf{Pre-training} We use Adamw optimizer \cite{ref39} to train the network with the initial learning rate 0.0001. The learning rate increases linearly for the first 10 epochs and then decays with a cosine schedule. We train the pre-training model with the batch size 64 and 200 epochs, and the whole pre-training is implemented on NVIDIA A100. For the exponential moving average weight $\lambda$ of the target network, the starting value is set to 0.996 and then gradually increases to 1. The dimension $K$ of the final features used to calculate the loss is set to 512. When cropping the global point cloud, the crop ratios $\gamma_{g1}$, $\gamma_{g2}$ are set to 0.7 and 1.0, respectively, and the number of crops $I$ is 2. When cropping local point clouds, the crop ratios $\gamma_{l1}$, $\gamma_{l2}$ are set to 0.2 and 0.5, respectively, and the number of crops $J$ is 8. Additionally, we use the FPS sample half of the original point cloud as different resolution point clouds and add them to local point cloud set. The number of different resolution point clouds is 2. 

\textbf{Classification} We use a fully connected MLP network that combines ReLU, BN, and Dropout operations as the classification head. The SGD is used as the optimizer to fine tune the classification network with cosine schedule. We set the batch size to 32.

\textbf{Segmentation} Different from the classification task, the segmentation task needs to predict  pre-point labels. We first select multiple stage features of network, including the initial input feature of standard transformer and the output features of layer 3 and layer 7. We cascade the features of these different layers, and then use the point feature propagation in PointNet++ to propagate the features of the 256 down sampled points to the 2048 raw input points. Finally, MLP is used to map the features to the segmentation label space. Our batch size is 16 with a learning rate initialized to 0.0002 and decayed via the cosine schedule. We use the Adamw optimizer to train the segmentation network.

\subsection{Dataset}
In the experiments of this paper, four datasets (ShapeNet \cite{ref33}, ModelNet40 \cite{ref31}, SacnObjectNN \cite{a25}, and ShapeNetPart \cite{ref34}) are used.

\textbf{ShapeNet} contains 57448 CAD models, with a total of 55 categories. For the acquisition of point cloud data, we follow the processing method of Yang et al., and sample 2048 points from each CAD model surface. We use ShapeNet dataset as pre-training dataset. In the pre-training stage, we use the farthest point sampling algorithm to select 64 group center points, and divide 2048 points into 64 groups, where each group contains 32 points.

\textbf{ModelNet40} contains 12,331 handmade CAD models of from 40 categories and is widely used for point cloud classification tasks. We follow Yu et al. to sample 8192 points from each CAD model surface. According to the official split, 9,843 are used for training and 2,468 for testing. Following the work of Yu et al. \cite{a20}, we generated a Fewshot-ModelNet40 dataset based on ModelNet40. "M-way N-shot" represents the data under different settings, where M-way represents the number of categories selected for training, N-shot represents the number of samples for each category, and the number of samples used for testing is 20. M is selected from 5 and 10, and N is selected from 10 and 20.

\textbf{SacnObjectNN} is a 3D point cloud classification dataset derived from real-world scanned data. It contains 2902 point clouds from 15 categories. Due to the noise of occlusion, rotation and background, it is more difficult to classify. Following Yu et al. \cite{a20}, we selected three variant datasets to conduct experiments, including OBJ-BG, OBJ-ONLY, and PB-T50-RS.

\textbf{ShapeNetPart} contains 16811 objects from 16 categories. Each object consists of 2 to 6 parts with total of 50 distinct parts among all categories. Following Yu et al. \cite{a20}, we randomly select 2048 points as input.

\begin{table}\scriptsize
  \centering
  \caption{\textbf{Classification results with linear SVM on ModelNet40}. These models are trained in ShapeNet.}
  \setlength{\tabcolsep}{0.5cm}{
  \begin{tabular}{l  c  c  r}\\
      \hline Method & Year & Input & Accuracy \\
      \hline
      SPH \cite{Tmethod1} & 2003 & voxel & $68.2 \%$ \\ 
      LFD \cite{Tmethod2} & 2003 & view & $75.5 \%$ \\ 
      T-L \cite{Tmethod3} & 2016 & view & $74.4 \%$ \\ 
      VConv-DAE \cite{Tmethod4} & 2016 & voxel & $75.5 \%$ \\ 
      3D-GAN \cite{method1} & 2016 & voxel & $83.3 \%$ \\
      Latent-GAN \cite{method2} & 2018 & point & $85.7 \%$ \\
      MRTNet \cite{method3} & 2018 & point & $86.4 \%$ \\
      SO-Net \cite{method4} & 2018 & point & $87.3 \%$ \\
      FoldingNet \cite{method5} & 2018 & point & $88.4 \%$ \\
      MAP-VAE \cite{method6} & 2019 & point & $88.4 \%$ \\
      VIP-GAN \cite{method7} & 2019 & view & $90.2 \%$ \\
      3D-PointCapsNet \cite{method8} & 2019 & point & $88.9 \%$ \\
      Jigsaw3D \cite{method9} & 2019 & point & $90.6 \%$ \\
      Rotation3D \cite{method10} & 2020 & point & $90.7 \%$ \\
      CMCV \cite{Tmethod10} & 2021 & point & $89.8 \%$ \\
      MID-FC \cite{Tmethod8} & 2021 & point & $90.3 \%$ \\
      GSIR \cite{Tmethod9} & 2021 & point & $90.4 \%$ \\
      PSG-Net \cite{Tmethod11} & 2021 & point & $90.9 \%$ \\
      STRL \cite{method11} & 2021 & point & $90.9 \%$ \\
      ParAE \cite{Tmethod5} & 2021 & point & $91.6 \%$ \\
      Point-BERT \cite{Tmethod6} & 2022 & point & $88.6 \%$ \\
      CrossPoint \cite{Tmethod7} & 2022 & point & $91.2 \%$ \\
      \textbf{POS-BERT (our)} & 2022 & point & $\mathbf{92.1\%}$ \\
      \hline
  \end{tabular}}
  \label{tab1}
\end{table}
\begin{figure*}[htp]
  \centering
  \includegraphics[width=\textwidth/2]{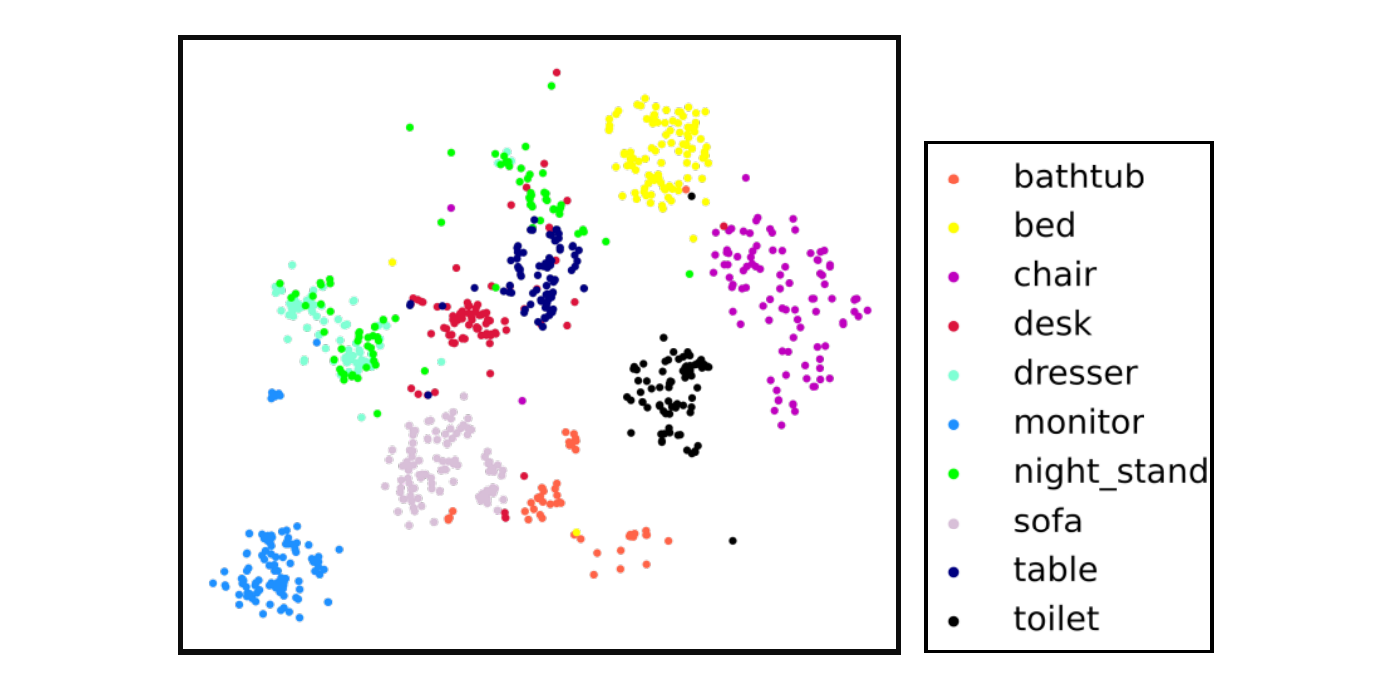} 
  \caption{Visualization of self-supervised features on ModelNet40.}
  \label{fig4}
\end{figure*}

\section{Experiment}
\subsection{Linear SVM Classification}
Linear SVM classification task has become a classic task to evaluate self-supervised point cloud representation learning. This experiment was designed to directly verify that our POS-BERT has learned better representation. To make a fair comparison with previous studies, we followed the common settings used in previous work \cite{a14, a15, a16, a19}, pre-trained the model on ShapeNet and tested it on the ModelNet40. We used our pre-training model to extract the features of each point cloud, then trained a simple linear Support Vector Machine (SVM) on the training set of ModelNet40, and finally tested the SVM on the ModelNet40 test set. We compared a series of competitive methods, including handcrafted descriptor methods, generation-based method, contrastive learning method, and the method based on mask patch modeling. The results of all methods are summarized in Tab.\ref{tab1}. The results of the comparison methods we reported adopt the best results in the original papers. As shown in Tab.\ref{tab1}, our method outperforms all other methods by a large margin, including the latest method CrossPoint based on contrastive learning and ParAE based on generation model. More importantly, it can surpass Point-BERT, which is also based on MPM paradigm, by 3.5\%. This result fully shows that our Momentum Encoder can provide more meaningful supervision representation for masked patches. Finally, it is worth mentioning that our linear classification results exceed some supervised point cloud networks, such as PointNet (89.7\%) and PointNet++ (91.9\%). For a more intuitive understanding of the performance of our model, we use t-SNE to map the self-supervised learn features to a 2D space, as shown in Fig.\ref{fig4}. It can be observed that different categories are separated from each other. These experimental results demonstrate that our method can learn a better representation.

\subsection{Downstream Tasks}
\textbf{3D Object Classification on Synthetic Data} 
To test whether POS-BERT can help boost downstream tasks. We first performed fine-tuning experiments on point cloud classification tasks using a pre-training model. Here, \textbf{From scratch} stands for training the model on ModelNet40 from randomly initialized network and \textbf{Pretrain} stands for pre-training the model on ShapeNet and then fine-tune the network on ModelNet40. We fine-tuned the classification network weights using different initialization methods on ModelNet40, and the final classification results were summarized in Tab.\ref{tab2}. Tab.\ref{tab2} shows that the original transformer's accuracy in point cloud classification task is just 91.4 percent. The transformer's classification accuracy was greatly increased to 93.56 percent using our pre-training weights to initialize the network. To achieve a fair comparison with Point-BERT, we also use voting strategy during the test, and the voting results are annotated with *. By comparison, we can see that our method outperforms OcCo and Point-BERT without voting by 1.4\% and 0.4\%, respectively. When using the voting strategy, even if the accuracy is already high, our method is slightly better than Point-BERT.

\begin{table}\scriptsize
  \centering
  \caption{\textbf{Shape classification results fine-tuned on ModelNet40.} We report the classification accuracy (\%).}
  \setlength{\tabcolsep}{0.5cm}{
\begin{tabular}{clcc}
  \hline Category & \multicolumn{1}{c}{ Method } & Input & Acc(\%) \\
  \hline \multirow{11}{*}{\textbf{From scratch}} 
  & PointNet  \cite{ref19} & point & $89.2$ \\
  & PointNet++ \cite{ref12} & point & $90.5$ \\
  & SO-Net \cite{method4} & point & $92.5$ \\
  & PointCNN \cite{ref20} & point & $92.2$ \\
  & DGCNN \cite{method13} & point & $92.9$ \\
  & DensePoint \cite{ref45} & point & $92.8$ \\
  & RSCNN \cite{ref46} & point & $92.9$ \\
  & PTC \cite{ref35} & point & $93.2$ \\
  & PointTransformer \cite{ref47} & point+nor & $93.7$ \\
  & NPTC \cite{ref35} & point & $91.0$ \\
  & Transformer \cite{ref49} & point & $91.4$ \\
  \hline \multirow{5}{*}{\textbf{Pretrain}} 
  & Transformer-OcCo \cite{ref50} & point & $92.10$ \\
  & Point-BERT \cite{ref49} & point & $93.16$ \\
  & \textbf{POS-BERT} & point & $\mathbf{93.56}$ \\
  \cline{2-4}
  & Point-BERT* \cite{ref49} & point & $93.76$ \\
  & \textbf{POS-BERT}* & point & $\mathbf{93.80}$ \\
  \hline
  \end{tabular}}
  \label{tab2}
\end{table}

\textbf{Few-shot Classification} 
To demonstrate that our pre-training model can learn quickly from few-shot samples, we conduct experiment on the Few-shot ModelNet40 dataset. We experimented with four different settings, including, "5-way 10-shot", "5-way 20-shot", "10-way 10-shot" and "10-way 20-shot", way represents the number of categories and shot represents the number of samples per category. During the test, 20 samples not in the training set were selected for evaluation. We conducted 10 independent experiments under each different setting, and reported the mean and variance of 10 experiments. We compared with the current SOTA methods OcCo and Point-BERT, and the results are summarized in Tab.\ref{tab3}. Our approach produces the best results on the Few-shot Classification task. Compared with baseline, the mean was increased by 8.6\%, 3.7\%, 8\%, and 5.5\%, respectively. The variance is almost halved. Compared with point-Bert, the mean increased by 1.8\%, 0.8\%, 1.6\% and 2.2\% respectively, and the variance was smaller. This completely demonstrates that POS-BERT has learned a universal representation suitable for quick knowledge transfer with limited data.

\begin{table}\scriptsize
  \centering
  \caption{\textbf{Few-shot classification results on ModelNet40.} We report the average accuracy (\%) as well as the standard deviation over 10 independent experiments.}
  \small
\begin{tabular}{lcccc}
  \hline & \multicolumn{2}{c}{ \textbf{5-way} } &\multicolumn{2}{c}{ \textbf{10-way} } \\
  \cline { 2 - 3 } \cline { 4 - 5 } & 10-shot & 20-shot &10-shot & 20-shot \\
  \hline DGCNN-rand \cite{ref50} & $31.6 \pm 2.8$ & $40.8 \pm 4.6$ &$19.9 \pm 2.1$ & $16.9 \pm 1.5$ \\
  DGCNN-OcCo \cite{ref50} & $90.6 \pm 2.8$ & $92.5 \pm 1.9$ & $82.9 \pm 1.3$ & $86.5 \pm 2.2$ \\
  \hline
  Transformer-rand \cite{ref49} & $87.8 \pm 5.2$ & $93.3 \pm 4.3$ & $84.6 \pm 5.5$ & $89.4 \pm 6.3$ \\
  Transformer-OcCo \cite{ref50} & $94.0 \pm 3.6$ & $95.9 \pm 2.3$ & $89.4 \pm 5.1$ & $92.4 \pm 4.6$ \\
  Point-BERT \cite{ref49} & $9 4 . 6 \pm 3 . 1$ & $9 6 . 3 \pm 2 . 7$ & $9 1 . 0 \pm 5 . 4$ & $9 2 . 7 \pm 5 . 1$ \\
  \textbf{POS-BERT} & $\mathbf{96.4 \pm 1.9}$ & $\mathbf{97.0 \pm 2.2}$ & $\mathbf{92.6 \pm 4.0}$ & $\mathbf{94.9 \pm 2.9}$ \\
  \hline
  \end{tabular}
  \label{tab3}
\end{table}

\textbf{3D Object Classification on Real-world Data} 
In this experiment, we aim to explore whether the knowledge POS-BERT learns from ShapNet can be transferred to real-world data. We conduct experiments on three variants of ScanObjectNN \cite{a25} dataset, including OBJ-BG, OBJ-ONLY, and PB-T50-RS. We compare to several methods, including supervised methods using specific point cloud networks: PointNet, BGA-PN++, SimpleView, et al., as well as pre-training methods: OcCo, Point-BERT. The experimental results are summarized in Tab.\ref{tab4}. It can be found from the table that our method obtains the best results. With OBG-BG and OBJ-ONLY, we have surpassed Point-BERT by 3.45\% and 2.76\%, respectively. We also outperform Point-BERT with the PB-T50-RS settings. The results of the experiments suggest that the knowledge learned by POS-BERT can easily transfer into real-world data.

\begin{table}\scriptsize
  \centering
  \caption{\textbf{Classification results on the ScanObjectNN dataset.} We report the accuracy (\%) of three different settings.}
  \setlength{\tabcolsep}{0.5cm}{
  \begin{tabular}{lccc}
    \hline Methods & OBJ-BG & OBJ-ONLY & PB-T50-RS \\
    \hline PointNet \cite{ref19} & $73.3$ & $79.2$ & $68.0$ \\
    SpiderCNN \cite{ref51} & $77.1$ & $79.5$ & $73.7$ \\
    PointNet++ \cite{method12} & $82.3$ & $84.3$ & $77.9$ \\
    PointCNN \cite{ref20} & $86.1$ & $85.5$ & $78.5$ \\
    DGCNN \cite{method13} & $82.8$ & $86.2$ & $78.1$ \\
    BGA-DGCNN \cite{ref52} & $-$ & $-$ & $79.7$ \\
    BGA-PN++ \cite{ref52} & $-$ & $-$ & $80.2$ \\
    SimpleView \cite{ref53} & $-$ & $-$ & $80.5$ \\
    \hline Transformer \cite{ref49} & $79.86$ & $80.55$ & $77.24$ \\
    Transformer-OcCo \cite{ref50} & $84.85$ & $85.54$ & $78.79$ \\
    Point-BERT \cite{ref49} & $87.43$ & $88.12$ & $83.07$ \\
    \textbf{POS-BERT} & $\mathbf{90.88}$ & $\mathbf{90.88}$ & $\mathbf{83.21}$ \\
    \hline
    \end{tabular}}
  \label{tab4}
\end{table}

\textbf{Part Segmentation} 
In this section, we explore how the pre-training model performs in the pre-point classification. We experimented on ShapeNetPart, a benchmark dataset commonly used in point cloud segmentation tasks. Compared with the classification task, the segmentation task needs to obtain the label of each point intensively. We compare it with the commonly used point cloud analysis networks and the most advanced self-supervised methods. The mean Intersection Over Union (mIOU) metric of various methods is reported in Tab.\ref{tab5}. From the table, our method is significantly better than the most advanced method Point-BERT on $\mathrm{mIoU}_{C}$. From a category perspective, we have exceeded other methods in most categories. These results show that our methods can also learn to distinguish details very well.

\begin{table}\scriptsize
  \centering
  \tiny
  \caption{\textbf{Part segmentation results on the ShapeNetPart dataset.} We report the
  mean IoU across all instances $\mathrm{mIoU}_{I}$ (\%), as well as the IoU (\%) for each categories and mean IoU across all categories $\mathrm{mIoU}_{C}$ (\%).}
  \begin{tabular}{lcc|ccccccccc}
    \hline \multirow{2}{*}{Methods} & \multirow{2}{*}{$\mathrm{mIoU}_{C}$} & \multirow{2}{*}{$\mathrm{mIoU}_{I}$} & aero & bag & cap & car & chair & e-phone & guitar & knife \\
    & &  & lamp & laptop & motor & mug & pistol & rocket & skateboard & table \\
    \hline \multirow{2}{*}{PointNet \cite{ref19}} & \multirow{2}{*}{$80.4$} & \multirow{2}{*}{$83.7$} & $83.4$ & $78.7$ & $82.5$ & $74.9$ & $89.6$ & $73.0$ & $91.5$ & $85.9$ \\
    & &  & $80.8$ & $95.3$ & $65.2$ & $93.0$ & $81.2$ & $57.9$ & $72.8$ & $80.6$ \\
    \cline{4-11}
    \multirow{2}{*}{PointNet++ \cite{method12}} & \multirow{2}{*}{$81.9$} & \multirow{2}{*}{$85.1$} & $82.4$ & $79.0$ & $87.7$ & $77.3$ & $90.8$ & $71.8$ & $91.0$ & $85.9$ \\
    & &  & $83.7$ & $95.3$ & $71.6$ & $94.1$ & $81.3$ & $58.7$ & $76.4$ & $82.6$ \\
    \cline{4-11}
    \multirow{2}{*}{DGCNN \cite{method13}} & \multirow{2}{*}{$82.3$} & \multirow{2}{*}{$85.2$} & $84.0$ & $83.4$ & $86.7$ & $77.8$ & $90.6$ & $74.7$ & $91.2$ & $87.5$ \\ 
    & &  & $82.8$ & $95.7$ & $66.3$ & $94.9$ & $81.1$ & $63.5$ & $74.5$ & $82.6$ \\
    \cline{4-11}
    \hline \multirow{2}{*}{Transformer \cite{ref49}} & \multirow{2}{*}{$83.4$} & \multirow{2}{*}{$85.1$} & $82.9$ & $85.4$ & $87.7$ & $78.8$ & $90.5$ & $80.8$ & $91.1$ & $87.7$ \\
    & &  & $85.3$ & $95.6$ & $73.9$ & $94.9$ & $83.5$ & $61.2$ & $74.9$ & $80.6$ \\
    \cline{4-11}
    \multirow{2}{*}{Transformer-OcCo \cite{ref50}} & \multirow{2}{*}{$83.4$} & \multirow{2}{*}{$85.1$} & $83.3$ & $85.2$ & $88.3$ & $79.9$ & $90.7$ & $74.1$ & $91.9$ & $87.6$ \\
    & &  & $84.7$ & $95.4$ & $75.5$ & $94.4$ & $84.1$ & $63.1$ & $75.7$ & $80.8$ \\
    \cline{4-11}
    \multirow{2}{*}{Point-BERT \cite{ref49}} & \multirow{2}{*}{$84.1$} & \multirow{2}{*}{$85.6$} & $84.3$ & $84.8$ & $88.0$ & $79.8$ & $91.0$ & $81.7$ & $91.6$ & $87.9$ \\
    & &  & $85.2$ & $95.6$ & $75.6$ & $94.7$ & $84.3$ & $63.4$ & $76.3$ & $81.5$ \\
    \cline{4-11}
    \multirow{2}{*}{\textbf{POS-BERT}} & \multirow{2}{*}{$\mathbf{84.2}$}  & \multirow{2}{*}{$\mathbf{86.0}$} & $84.9$ & $86.4$ & $87.4$ & $81.0$ & $91.3$ & $78.4$ & $92.0$ & $88.2$ \\
    & &  & $85.0$ & $95.5$ & $76.0$ & $94.9$ & $84.7$ & $63.9$ & $75.9$ & $82.1$ \\
    \hline
  \end{tabular}
  \label{tab5}
\end{table}

\subsection{Ablation study} 
To demonstrate the effectiveness of our key modules, we conducted ablation study on the ModelNet40 Linear SVM classification task. We have designed four variants. The first variant uses a randomly initialized Transformer network to extract features directly without any pre-training, and then classifies them using SVM, which is defined as POS-BERT-Var1. The second variant, defined as POS-BERT-Var2, uses only masking patch modeling's pretext task for pre-training. The third variant uses the randomly initialized momentum encoder as the tokenizer to pre-training, which is defined as POS-BERT-Var3. The fourth variant, which uses only contrastive loss to train the point cloud transformer, is defined as POS-BERT-Var4. The results are summarized in Tab.\ref{tab6}. From the table we can see that a fixed Momentum Encoder does not help the network train well. Pre-training with masking patch modeling alone is difficult to obtain high-level semantic information. The best results are obtained when masking patch modeling and contrastive learning work together.

\begin{table}\scriptsize
  \centering
  \tiny
  \caption{\textbf{Ablation study.} .}
  \setlength{\tabcolsep}{0.5cm}{
  \begin{tabular}{c|ccc|c}
    \hline Model Name & MPM & GFC & Momentum Encoder & Acc(\%). \\
    \hline POS-BERT-Var1 & & & & $53.61$ \\
    POS-BERT-Var2 & $\checkmark$ & & & $80.43$ \\
    POS-BERT-Var3 & $\checkmark$ & $\checkmark$ & & $79.05$ \\
    POS-BERT-Var4 & & $\checkmark$ & $\checkmark$ & $91.29$ \\
    POS-BERT & $\checkmark$ & $\checkmark$ & $\checkmark$ & $92.14$ \\ 
    \hline
  \end{tabular}}
\label{tab6}
\end{table}

\section{Conclusion}
In this paper, we propose a one-stage point cloud pre-training method POS-BERT, which is simple, flexible and efficient. It uses momentum encoder as tokenizer to provide supervision for mask patch model pretext tasks, and joint training of momentum encoder and MPM tasks greatly simplifies the training steps and saves training costs. Experiments show that our method has the best ability to extract high-level semantic information in the Linear SVM classification task, and it improves significantly compared with Point-BERT. At the same time, many downstream tasks, including 3D object classification, few-shot classification, part segmentation, have achieved state-of-the-art performance.

\bibliographystyle{IEEEtran}
\bibliography{egbib}

\end{document}